\documentclass[11pt,a4paper]{article}
\usepackage[hyperref]{acl2021}
\usepackage{times}
\usepackage{latexsym}

\usepackage{microtype}

\aclfinalcopy 

\usepackage{latexsym}
\usepackage{amsmath,amssymb,bm}
\usepackage{graphicx}
\usepackage{caption}
\usepackage{algorithm}
\usepackage{algorithmic}
\usepackage{gensymb}
\usepackage{booktabs,caption}
\usepackage[flushleft]{threeparttable}
\usepackage{lineno}
\usepackage{kmath}

\usepackage[normalem]{ulem}

\newcommand{\iobfg}{\textsc{iob$_2$}}

\newcommand{\potbb}{\textsc{s+bert+\iobfg}}
\newcommand{\pot}{\textsc{s$_\text{adv}$+bert+\iobfg}}

\newcommand{\ours}{\textsc{SPADE}$\varspadesuit$}
\newcommand{\ourss}{\textsc{SPADE$\varspadesuit^\dagger$}}

\newcommand{\oursb}{\ours\ w/o \caabb}
\newcommand{\oursabb}{\textsc{$\varspadesuit$}}
\newcommand{\oursbabb}{\textsc{$\varspadesuit$} - \caabb}

\newcommand{\ubns}{\textsc{UB-no-ser}}

\newcommand{\ubsiob}{\textsc{UB-flat}}

\newcommand{\caabb}{\textsc{tca}}

\newcommand{\cord}{\textsc{CORD}}
\newcommand{\cordabb}{\textsc{co}}
\newcommand{\cordd}{\textsc{CORD++}}
\newcommand{\corddl}{\textsc{CORD+}}
\newcommand{\corddabb}{\textsc{co++}}
\newcommand{\corddlabb}{\textsc{co+}}
\newcommand{\cordtc}{\textsc{CORD-M}}
\newcommand{\cordtcabb}{\textsc{co-m}}
\newcommand{\funsdabb}{\textsc{fu}}
\newcommand{\oururl}{\texttt{https://github.com/clovaai/spade}}

\newcommand{\receipti}{\textsc{Receipt-idn}} 
\newcommand{\receiptiabb}{\textsc{ri}} 

\newcommand{\ncabb}{\textsc{nc}} 
\newcommand{\nc}{\textsc{namecard}} 

\newcommand{\invoice}{\textsc{Invoice}} 
\newcommand{\invoiceabb}{\textsc{inv}} 

\DeclareMathOperator*{\argmax}{arg\,max}

\usepackage{makecell}

\usepackage{color}

\graphicspath{{./figures/}}

\title{Spatial Dependency Parsing for \\Semi-Structured Document Information Extraction}

\author{Wonseok Hwang \quad Jinyeong Yim \quad Seunghyun Park \quad Sohee Yang \quad Minjoon Seo \\
  Clova AI, NAVER Corp. \\
  \texttt{\{wonseok.hwang,jinyeong.yim,seung.park,sh.yang,minjoon.seo\}}\\\texttt{@navercorp.com}
}

\date{}

\begin{document}
\maketitle
\begin{abstract}
Information Extraction (IE) for semi-structured document images is often approached as a sequence tagging problem by classifying each recognized input token into one of the IOB (Inside, Outside, and Beginning) categories. 
However, such problem setup has two inherent limitations that (1) it cannot easily handle complex spatial relationships and (2) it is not suitable for highly structured information, which are nevertheless frequently observed in real-world document images. 
To tackle these issues, we first formulate the IE task as \emph{spatial dependency parsing} problem that focuses on the relationship among text tokens in the documents.
Under this setup, we then propose \ours\ (SPAtial DEpendency parser) that models highly complex spatial relationships and an arbitrary number of information layers in the documents in an end-to-end manner.
We evaluate it on various kinds of documents such as receipts, name cards, forms, and invoices, and show that it achieves a similar or better performance compared to strong baselines including BERT-based IOB taggger. 

\vspace{-2mm}
\end{abstract}
\begin{figure*}[t!]
\centering
\includegraphics[width=1\textwidth]{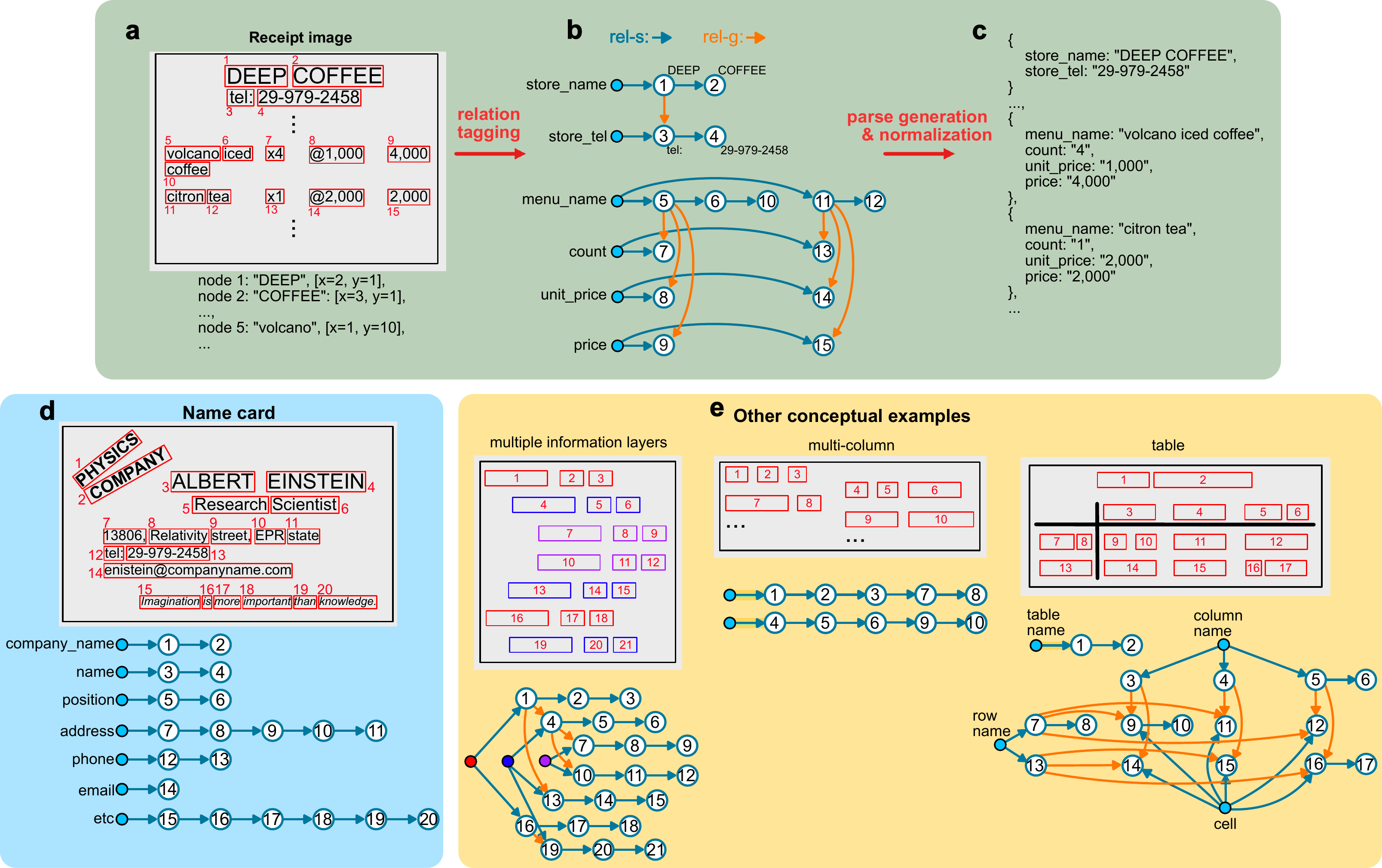}
\caption{ 
The illustration of spatial dependency parsing problem. Receipt parsing is explained in detail with three subfigures: (a) first, text tokens and their coordinates are extracted from OCR; (b) next, the relations between tokens are classified into two types: \texttt{rel-s} for serialization and information type (field) classification, and \texttt{rel-g} for inter-grouping between fields (the numbers inside of circles in (b) indicates the box numbers in (a)); (c) the final parse is generated by decoding the graph.
(d) A sample name card and its spatial dependency parse.
(e) Other conceptual examples showing the versatility of the spatial dependency parsing approach for document IE.
}
\label{fig_paradigm}
\vspace{-4mm}
\end{figure*}

\section{Introduction}
\noindent Document information extraction (IE) is the task of mapping each document to a structured form that is consistent with the target ontology (e.g., database schema), which has become an increasingly important task in both research community and industry.
In this paper, we are particularly interested in information extraction from real-world, semi-structured document images, such as invoices, receipts, and name cards, where we assume Optical Character Recognition (OCR, i.e. detecting the locations of the text tokens if the input is an image) has been already applied.
Previous approaches for semi-structured document IE often assume as if the input is a one-dimensional sequence and formulate the task as an IOB (Inside Outside Beginning) tagging problem.
In this setup, the tokens in the document (either obtained through an OCR engine or trivially parsed from a web page or pdf) are first serialized, and then an independent tagging model classifies each of the flattened lists into one of the pre-defined IOB categories~\cite{ramshaw1995bio,palm2017cloundscan}.
While effective for relatively simple documents, their broader application in the real world is still challenging because (1) semi-structured documents often exhibit a complex layout where the serialization algorithm is non-trivial, and (2) sequence tagging is inherently not effective for encoding multi-layer hierarchical information such as the menu tree in receipts (Fig.~\ref{fig_paradigm}c).

To overcome these limitations, we propose \ours\ (SPAtial DEpendency parser), an end-to-end, serializer-free model that is capable of extracting hierarchical information from complex documents.
Rather than explicitly dividing the original problem into two independent subtasks of serialization and tagging, our model tackles the problem in an end-to-end manner by creating a directed relation graph of the tokens in the document (Fig.~\ref{fig_paradigm}).
In contrast to traditional dependency parsing, which parses the dependency structure in purely (one-dimensional) linguistic space, our approach leverages both linguistic and (two-dimensional) \emph{spatial} information to parse the dependency.

We evaluate \ours\ on eight document IE datasets created from real-world document images, including invoices, name cards, forms, and receipts, with the varying complexity of information structure.
In all of the datasets, our model shows a similar or better accuracy than strong baselines including BERT-based IOB taggers, and particularly outstands in documents with complex layouts (Table \ref{tbl_results_f1}).
These results demonstrate the effectiveness of our end-to-end, graph-based paradigm over the existing sequential tagging approaches.

In short, our contributions are threefold. (1) We present a novel view that information extraction for semi-structured documents can be formulated as a dependency parsing problem in two-dimensional space. (2) We propose \ours\ for spatial dependency parsing, which is capable of efficiently constructing a directed semantic graph of text tokens in semi-structured documents.\footnote{\oururl} (3) \ours\ achieves a similar or better accuracy than the previous state of the art or strong BERT-based baselines in eight document IE datasets.

\section{Related Work}
The recent surge of interest in automatic information extraction from semi-structued documents are well reflected in their increased number of publication record from both research community and industry \citep{katti2018chargrid,qian2019graphie,liu2019graph,zhao2019cutie,denk2019arXiv,hwang2019pot,park2019cord,xu2019_layoutLM,jaume2019funsd,zhong2019pubtabnet,rausch2019docparser,yu2020_pick,wei2020,majumder2020repr,lockard2020_zeroshotceres,garncarek2020lambert,lin2020_freedom,xu2020layoutlmv2,powalski2021tilt,wang2021vies,hong2021bros,hwang2021wyvern}.
Below, we summarize some of closely related works published before the major development of \ours.

\paragraph{Serialized IE} Previous semi-structured document information extraction (IE) methods often require the input text boxes (obtained from OCR) to be serialized into a single flat sequence. 
\citet{hwang2019pot} and \citet{denk2019arXiv} combine a manually engineered text serializer that turn the OCR text boxes into a sequence and a Transformer-based encoder, BERT \cite{devlinBERT2018}, that performs IOB tagging on the sequence or semantic segmentation from images. 
In contrast to \ours, these models rely on the serialization of the tokens and thus it is difficult to flexibly apply them to documents with complex layouts such as multi-column or distorted documents. 
\citet{xu2019_layoutLM} propose LayoutLM that jointly embeds the image segments, text tokens, and positions of the tokens in an image to make a pretrained model for document understanding. However, LayoutLM still requires a careful serialization of the tokens as it relies on the position embeddings of BERT. Also, it is only evaluated on classification for the downstream task.

\paragraph{Serializer-free IE} Existing serializer-free methods mostly extract flat key-value pairs, as they still formulate the task as tagging the text tokens. They fundamentally differ from \ours\ which generates a structured output that captures full information hierarchy represented in the document.
Chargrid~\citep{katti2018chargrid} performs semantic segmentation on invoice images to extract target key-value pairs. 
Although Chargrid uses additional ``bounding boxes'' for inter-grouping of certain fields, the application to the documents that have more than two information hierarchy levels is non-trivial. 
Also, when fields that belong to the same group are remotely located, the bounding boxes may need to be modified to have a more complex geometrical shape to avoid overlap between the boxes.

\paragraph{Graph-based IE} \citet{liu2019graph,qian2019graphie,wei2020,yu2020_pick} utilize a graph convolution network to contextualize the tokens in a document and a bidirectional LSTM with CRF to predict the IOB tags. 
However, the range of possible parse generations is limited as IOB tagging can be performed only within each OCR bounding box, ignoring inter-box relationship.
On the contrary, \ours\ predicts both the intra-box relationship and the inter-box relationship by constructing a dependency graph among the tokens.

\citet{lockard-etal-2019-openceres,lockard2020_zeroshotceres} also utilize a graph to extract semantic relation from semi-structrued web-page. The graph is constructed based on rules from ``structured html DOM'' and mainly used for information encoding. On the other hand \ours\ accepts ``unstructured text distributed in 2D'' and generates graphs as the result of decoding (in a data-driven way).

\begin{figure}[t]
\centering
\includegraphics[width=0.47\textwidth]{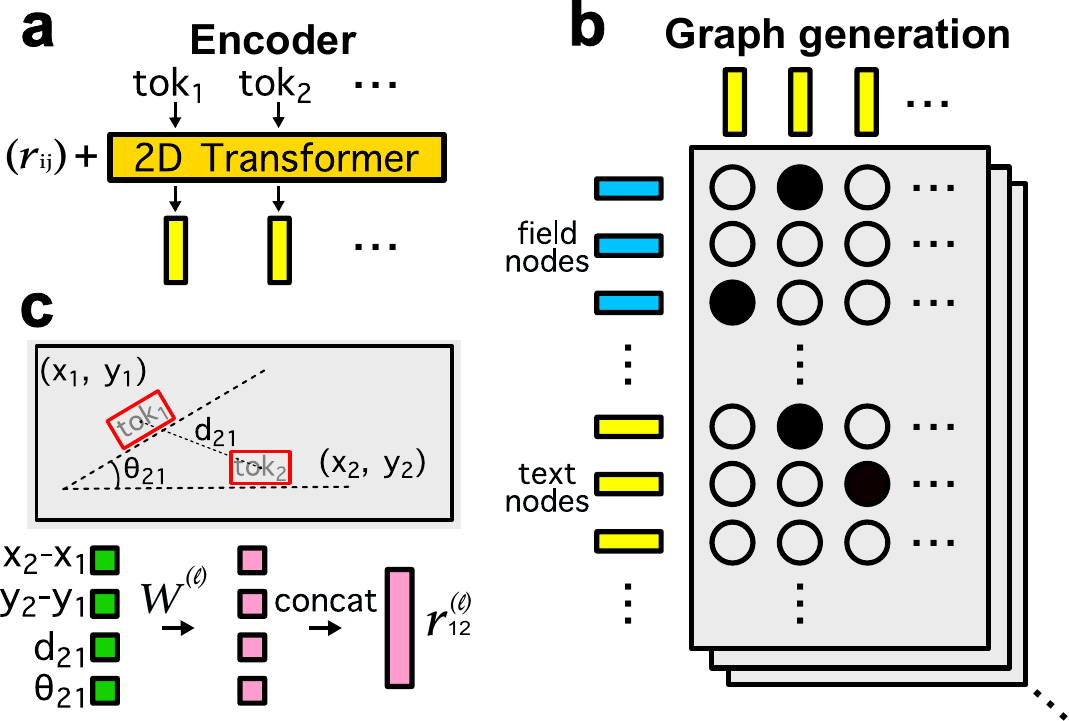}
\caption{
The illustration of \ours. (a) Spatial text encoder contextualizes tokens using their relative spatial vectors $\{r_{ij}\}$ (Eq. \ref{eqn:txl}). (b) The dependency graph is inferred by mapping the vector representations of each pair of tokens into a scalar value. The field embeddings are blue-colored. (c) At each encoder layer $\ell$, $r_{ij}^{\ell}$ is prepared by concatenating four embedding vectors: relative coordinates, distance, and angles embeddings. $W^{\ell}$ stands for a linear projection.
}
\label{fig_model}
\vspace{-4mm}
\end{figure}

\paragraph{Dependency parsing} 
Dependency parsing is the task of obtaining the syntactic or semantic structure of a sentence by defining the relationships between the words in the sentence~\citep{zettlemoyer2012learning, peng2017cross, dozat2018simpler_dependency}. The relations are often expressed as directed, labeled arcs. In our work, we view the problem of information extraction for semi-structured documents as a \emph{spatial} dependency parsing task such that two-dimensional spatial information is mainly considered. This setup enables \ours\ to flexibly handle documents with complicated layouts while representing the full information hierarchy.

\section{Problem definition}
In this section, we first describe the task of information extraction for semi-structured documents, and we briefly discuss how the task was approached in the past as a sequence tagging problem.
Then we formulate it as a spatial dependency parsing problem. In Section \ref{sec:model}, we show how we design our model for the newly formulated problem.

\subsection{Semi-structured document IE}\label{subsec:ie}
Document IE is often defined as the extraction of structured information (e.g. key-value pairs) in documents.
For semi-structured documents, the task becomes more challenging, mainly due to two factors: (1) complex spatial layout and (2) hierarchical information structure.
In the simplest case, both of the two factors are minimally present, where the text is strictly a linear sequence, and the desired output is simply a list of fields, similar to Named Entity Recognition (NER) task.
However, the problem becomes more difficult when at least one of the factors is significant.
In name cards, spatial relationship can be tricky;  Fig. \ref{fig_paradigm}d shows an example where a na\"ive left-to-right serialization would fail because the \texttt{company\_name} (``Physics Company'') 
is tilted. 
In receipts, their hierarchical information structure complicates the problem.
For example, in Fig. \ref{fig_paradigm}a), words ``volcano" (box 5), ``iced" (box 6), and ``coffee" (box 10) together form a single field \texttt{menu\_name}, and the field constitutes another group in the second hierarchical layer 
with the \texttt{count} field (box7), \texttt{unit\_price} field (box 8), and \texttt{price} field (box 9). 
Other conceptual examples are shown in Fig. \ref{fig_paradigm}e); documents that have triple information layers (left), multiple columns (middle), and a table (right).

\subsection{Previous formulation: Sequence tagging}\label{subsec:tag}
As mentioned, IOB sequence tagging is appropriate for document IE when the layout and the information structure are simple~\cite{ramshaw1995bio,lample2016neural_ner,chiu-nichols2016named,ma2020neural_seqtag}.
When one of the factors is present, however, one has to adopt an ad-hoc solution to detour the inherent limitation of IOB.

In the case of complex spatial relationship (e.g., name card), an advanced, dedicated serialization method can be considered.
However, it may require layout-specific manual engineering, which becomes more difficult for documents such as  name cards that exhibit diverse layouts.

In the case of complex information 
structure (e.g., receipt), one can consider augmenting each IOB tag with higher-layer information. 
For instance, in a typical IOB setting, the \texttt{menu\_name} field will require two tags, namely \texttt{menu\_name\_B} and \texttt{menu\_name\_I}. To model the second layer information (inter-grouping of fields), \texttt{menu\_name\_B} can be augmented into two, namely \texttt{B2\_menu\_name\_B}, \texttt{I2\_menu\_name\_B}, where \texttt{B2} and \texttt{I2} indicate the beginning and the inside of the hierarchy's second layer.
While effective for some applications, this method would not generalize well to an arbitrary depth as it requires more tags for each additional layer.

\subsection{Our formulation: Spatial dependency parsing}\label{subsec:dep}

To better model spatial relationship and hierarchical information structure in semi-structured documents,
we formulate the IE problem as ``spatial dependency parsing'' task by constructing a dependency graph with tokens and fields as the graph nodes (node per token and field type).
This is demonstrated in Fig.~\ref{fig_paradigm}, where empty blue circles are text nodes, and filled blue circles are field nodes.

Although the spatial layout of semi-structured documents is diverse, it can be considered as the realization of mainly two abstract properties between each pair of nodes, (1) \texttt{rel-s} for the ordering and grouping of tokens belonging to the same information category (blue arrows in Fig. \ref{fig_paradigm}b), and (2)  \texttt{rel-g} for the inter-group relation between grouped tokens or groups (orange arrows in the same figure).
Connecting a field node to a text node indicates that the text is classified into the field.
For example, ``volcano iced coffee'' in Fig.~\ref{fig_paradigm}a) is classified as a menu name by being attached to the \texttt{menu\_name} field node with blue arrows, and it is connected with ``x4'', ``@1,000'', and ``4,000'' with orange arrows to indicate the hierarchical information among the groups. 
The dependency graphs of name cards and other conceptual examples are also shown in Fig. \ref{fig_paradigm}d and e.

\section{Model}\label{sec:model}
To perform the spatial dependency parsing task introduced in the previous section in an end-to-end fashion, we propose \ours\ that consists of (1) spatial text encoder, (2) graph generator, and (3) graph decoder. Spatial text encoder and graph generator are trained jointly. Graph decoder is a deterministic function (without trainable parameters) that maps the graph to a valid parse of the output structure.

\subsection{Spatial text encoder}\label{subsec:enc}

Spatial text encoder is based on 2D Transformer architecture. Unlike the original Transformer \citep{vaswani2017transformer}, there is no order among the input tokens, making the model invariant under the permutation of the input tokens.
Inspired by Transformer XL \citep{dai2019transformer-XL}, the attention weights (between each key and query vector) is computed by
\begin{equation}\label{eqn:txl}
    \begin{aligned}
        q_i^T k_j + q_i^T r_{ij} + (b_i^{key})^T k_j + (b_i^{rel})^T r_{ij}
    \end{aligned}
\end{equation}
where $q_i$ is the query vector of the $i$-th input token, $k_j$ is the key vector of the $j$-th input token, $r_{ij}$ is the relative spatial vector of the $j$-th token with respect to the $i$-th token, and $b_i^{key|rel}$ is a bias vector. 
In (original) Transformer, only the first term of Equation~\ref{eqn:txl} is used.

The relative spatial vector $r_{ij}$ is constructed as follows (Fig. \ref{fig_model}c).
First, the relative coordinates between each pair of tokens are computed.\footnote{For example, if ``token1'' is at $(x_1=1, y_1=10)$ and ``token2'' is at $(x_2 = 3, y_2=4)$, the relative coordinate of ``token2'' with respect to ``token1'' is $(x'_2, y'_2) = (3-1, 4-10)=(2, -6)$.}
Next, the coordinates are quantized into integers and embedded using \texttt{sin} and \texttt{cos} functions \citep{vaswani2017transformer}. The physical distance and the relative angle between each pair of the tokens are also embedded in a similar way.
Finally, the four embedding vectors are linearly projected (with a trainable projection matrix) and concatenated at each encoder layer. 

\subsection{Graph generator} \label{sec:graph_gen}
As discussed in Section \ref{subsec:dep} and shown in Fig.~\ref{fig_paradigm}, every token corresponds to a node and each pair of the nodes forms one of the two relations (or no relation):
(1) \texttt{rel-s} for serializing tokens within the same field, and (2) \texttt{rel-g} for inter-grouping between fields. 
The dependency graph can be represented by using a binary matrix $M^{(r)}$ for each relation type $r$ (Fig. \ref{fig_model}b) 
where $M_{ij}^{(r)} = 1$ if their exists a directed edge from the $i$-th token to the $j$-th token and $0$ otherwise.
Each $M^{(r)}$ consists of $n_{\text{field}} + n_{\text{text}}$ number of rows and $n_{\text{text}}$ number of columns where $n_{\text{field}}$ and $n_{\text{text}}$ represent the number of field types and the number of tokens, respectively.
The graph generation task now becomes predicting the binary matrix.

We obtain $M^{(r)}$ as follows. 
The probability that there exists a directed edge $i \xrightarrow{r} j$ is computed by
\begin{equation}
    \begin{aligned}
    h^{(r)}_i &=
    \begin{cases}
    u^{(\text{field})}_i, ~~ \text{for $i \leq n_{\text{field}}$}
    \\ \mathcal{W}^{(r)}_h v_i  ~~ \text{otherwise}
    \end{cases}
    \\ d^{(r)}_j &= \mathcal{W}^{(r)}_d v_j 
    \\ s_{0,ij}^{(r)} &= (h^{(r)}_i)^T \mathcal{W}^{(r)}_0 d^{(r)}_j
    \\ s_{1,ij}^{(r)} &= (h^{(r)}_i)^T \mathcal{W}^{(r)}_1 d^{(r)}_j
    \\ p_{ij}^{(r)} &= \frac{\exp({s_{1,ij}^{(r)}}) }{\exp({s_{0,ij}^{(r)}}) + \exp({s_{1,ij}^{(r)}}) },
    \end{aligned}
\end{equation}
where $u^{(\text{field})}_i$ represents the trainable embedding vector of the $i$-th field type node (filled blue circles in Fig. \ref{fig_paradigm}), 
$\{v_i\}$ is a set of vectors of contextualized tokens from the enoder,
$\mathcal{W}$ stands for affine transformation, $h$ is the embedding vector of the head token, and $d$ is that of the dependent token.

$M_{ij}^{(r)}$ is obtained by binarizing $p_{ij}^{(r)}$ as follows.
\begin{equation} \label{func:binarization}
    \begin{aligned}
     M_{ij}^{(r)}(p_{ij}^{(r)}) &= 
    \begin{cases}
    \text{\small{1 for ($r$=\texttt{s}, $i$ is field type node, $p_{ij}^{(r)} \geq p_{th}$)} }
    \\  \text{\small{1 for ($r$=\texttt{s}, $i$ is text node, $p_{ij}^{(r)} \geq p_{th}$,} }
    \\ ~~~~~~~~\text{\small{$j$ = $\argmax_k p_{ik}^{(r)}$)}}
    \\  \text{\small{1, for ($r$=\texttt{g}, $i$ is text node, $p_{ij}^{(r)} \geq p_{th}$)} }
    \\  \text{\small{0, otherwise.}}
    \end{cases}
    \end{aligned}
\end{equation}
The recall rate of edges can be controlled by varying the threshold value $p_{th}$. Here, we set $p_{th}=0.5$.

\paragraph{Tail collision avoidance algorithm}
Each node in spatial dependency graphs has a single incoming edge per relation except some special documents such as table (Fig. \ref{fig_paradigm}e). 
Based on this property, we apply the following simple yet powerful tail collision avoidance algorithm:
(1) at each tail node having multiple incoming edges, all edges are trimmed except the one with the highest linking probability; (2) at each head node of the trimmed edges, the new tail node is found by drawing the next probable edge whose probability is larger than $p_{th}$ and belongs to the top three; 
(3) go back to Step 1 and repeat the routine until the process becomes self-consistent or the max iteration limit is reached (set to 20 in this paper).
The algorithm prevents loops and token redundancy in parses.

\subsection{Graph decoder} \label{sec:pd}

We decode the generated graph into the final parse through the following three stages: (1) \texttt{SEEDING}, (2) \texttt{SERIALIZATION}, and (3) \texttt{GROUPING} (Table \ref{tbl_parse_gen}). In \texttt{SEEDING}, field type nodes (filled circles in Fig. \ref{fig_paradigm}) are linked to multiple text nodes (seeds) by \texttt{rel-s}. In \texttt{SERIALIZATION}, each seed node found in the previous stage generates a directed edge (\texttt{rel-s}) to the next text node (i.e. serialization) recursively until there is no further node to be linked. Finally, in \texttt{GROUPING}, the serialized texts are grouped iteratively, constructing information layers from the top to the bottom. The total number of iterations is equal to ``the number of information layers$-1$". To group texts using directed edges, we define a special representative field for each information layer. Then, the first token of the representative field generates directed edges to the first token of other fields that belong to the same group using \texttt{rel-g} (for example, \texttt{menu\_name} (``volcano iced coffee") in Fig. \ref{fig_paradigm}a) generates directed edges to other member fields (\texttt{count} (``x4"), \texttt{unit\_price} (``@1,000") and \texttt{price} (``4,000")).

The process generates an \emph{arborescence}\footnote{A directed graph in which, for a vertex u called the root and any other vertex v, there is exactly one directed path from u to v (Excerpted from Wikipedia)} for each field (\texttt{rel-s}) and group (\texttt{rel-g}). The resulting set of graphs has a one-to-one correspondence with the parse through detokenization. 
The use of beam search in \texttt{SERIALIZATION} does not introduce noticeable difference in \texttt{rel-s} probably due to the short decoding length of the graph (mostly less than 30).
The development of a more advanced decoding algorithm that generates globally optimal multiple arborescences remains as future work.

Although undirected edges can be employed for the inter-grouping of fields, the use of directed edges has the following merits:
(1) an arbitrary depth of information hierarchy can be described without increasing the number of relation types (Fig. \ref{fig_paradigm}e) under a unified framework and (2) a parse can be generated in a straightforward manner by iteratively selecting dependent nodes.

\begingroup
\setlength{\tabcolsep}{1.pt} 
\renewcommand{\arraystretch}{1} 
\begin{table}[h!]
\captionsetup{font=small} 
  \caption{A formal description of the parse decoding process. 
\texttt{s} and \texttt{g} stand for \texttt{rel-s} and \texttt{rel-g} respectively.
  }
  \label{tbl_parse_gen}
  \scriptsize
  \centering
  \begin{tabular}{lcl}
    \toprule
    Action & Input node  & Graph at time $t+1$  \\
    \midrule
    \midrule
    \texttt{INITIALIZATION} & & $G_{t=0} = \text{empty set} $  \\
    \midrule
     \texttt{SEEDING}($\mu$)  &  $\mu \in$ field nodes & $G^{(\text{seed})} = \{ \mu \xrightarrow{\texttt{s}} j | M_{\mu j}^{(\texttt{s})} = 1\} $ \\
     \midrule
      \texttt{SERIALIZATION}($i$) &  $i \in G^{(\text{seed})} \cup G_t$  &  $G_{t+1}=G_t \cup \{i \xrightarrow{\texttt{s}} j | M_{i j}^{(\texttt{s})} = 1\}$ \\
    \midrule
     \texttt{GROUPING}($i$) & \makecell{$i \in G^{(\text{seed})} \cup G_t$ \\ $i$ linked to representer fields}   &  $G_{t+1} = G_t \cup \{i \xrightarrow{\texttt{g}} j |  M_{i j}^{(\texttt{g})}=1\}  $  \\
    \midrule
     \texttt{MERGE} &  & $G_t = G_t \cup G^{(\text{seed})}$  \\
    \bottomrule
  \end{tabular}
  \vspace{-4mm}

\end{table}
\endgroup

\section{Experimental Setup} \label{sec:exp}

\subsection{Optical character recognition}
To extract the visually embedded texts from an image, we use our in-house OCR system that consists of CRAFT text detector~\citep{baek2019craft} and Comb.best text recognizer~\citep{baek2019wrong}. The OCR models are finetuned on each of the document IE datasets. The output tokens and their spatial information on the image are used as the inputs to \ours.

\subsection{Training}
We use 12 layers of 2D Transformer encoder (Section \ref{subsec:enc}). 
The parameters are initialized from \texttt{bert-multilingual} \citep{devlinBERT2018} \footnote{ https://github.com/huggingface/transformers}.
ADAM optimizer \citep{kingam2015adam} is used with the following learning rates: 1e-5 for the encoder, 1e-4 for the graph generator, and 2e-5 for \potbb\ and \pot. The decay rates are set to $\beta_1 = 0.9, \beta_2 = 0.999$. The batch size is chosen between 4 and 12. \ours\ is trained by using one to eight NVIDIA V100 or P40 GPUs for two to seven days, depending on the tasks.
The dev sets are used to pick the best model except FUNSD task in which the model is trained in two steps. First, the 25 examples from training set are sampled and used for a model validation. Next, the model is further trained using entire training set and stopped after 1000 epochs.
The training dataset is augmented by randomly rotating the text coordinates by a degree of -10$^{\circ}$ to +10$^{\circ}$, (2) by distorting the whole coordinates randomly using a trigonometric function, and (3) by randomly deleting or inserting a single token with 3.3\% probability each. Also, 1--2 random tokens from training is attached at the end of the text segments from OCR bounding box with 1.7\% probability each. In \nc\ task, the tokens are not augmented. 
The identical augmentation algorithm are applied to \potbb, \pot\ and \ours.

\subsection{Evaluation metric}
To evaluate the predicted parses that consist of hierarchically organized key-value pairs (e.g. Fig. \ref{fig_distortion}, Fig. \ref{fig_two_column}, \ref{fig_cord_examples1}, \ref{fig_cord_examples2} in Appendix) we use $F_1$ score based on exact match. First the group of key-value pairs between predictions and ground truth (gt) are matched based on their string edit distance. Each key-value pairs in the predicted parse is counted as true positive if same key-value pair exists within the corresponding group in gt. Otherwise it is counted as false positive. The unmatched key-value pairs in gt are counted as false negative. The accuracy of dependency parsing is evaluated by computing $F_1$ of predicted edges.
For FUNSD dataset, entity labeling and entity linking scores are computed following the original paper \citep{jaume2019funsd}. See Appendix \ref{sec:eval_detail} for more details. 

\subsection{Data statistics}
We summarize the data statistics in Table \ref{tbl_datasets}, \ref{tbl_datasets_full}. The property of each dataset and their collection process is described in Appendix \ref{sec:dataset_detail}.

\begingroup
\setlength{\tabcolsep}{0.9pt} 
\renewcommand{\arraystretch}{0.9} 
\begin{table}[h!]
\centering
\begin{threeparttable}

\captionsetup{font=small} 
  \caption{
  The dataset properties. 
  }
  \label{tbl_datasets}
  \tiny

  \centering

  \begin{tabular}{l@{\extracolsep{\fill}}cccccccc}

    \toprule
     Dataset & Lang.&Abbr. &\makecell{\# of field\\types} & \makecell{\# of examples\\(train:dev:test)}  & \makecell{\# of fields} & \makecell{Mean \# of\\ text nodes}  & Depth & \makecell{Layout\\complexity} \\
     \midrule
     \midrule
     \cord & IDN &\cordabb &30 & 800:100:100 & 13030 & 62.3& 2 & low \\
     \corddl & IDN & \corddlabb&\texttt{"} & \texttt{"} & \texttt{"} & 62.3& 2 & high  \\
     \cordd & IDN &\corddabb &\texttt{"} & \texttt{"} & \texttt{"} & 62.3& 2 & high  \\
     \cordtc & IDN &\cordtcabb&\texttt{"} &400:50:50 & \texttt{"}& 124.6& 3 & low \\
     \receipti & IDN &\receiptiabb& 50 & 9508:458:450 & 209728 & 209 & 2 & low \\
     \nc & JPN &\ncabb& 12 & 22076:256:100 & 231528 & 19.4 & 1 & high  \\
     \invoice & JPN&\invoiceabb & 62 & 896:79:83 & 37115 & 412 & 2 & high \\
     FUNSD$^{a}$ & ENG&\funsdabb& 4 & 149:50 & 9743 & 179& 3 & high  \\
    \bottomrule
  \end{tabular}
\begin{tablenotes}
\scriptsize
\item[a]{The statistics are from \citet{jaume2019funsd}.} 
\end{tablenotes}
\end{threeparttable}
\vspace{-4mm}
\end{table}
\endgroup

\section{Experimental Results}

\begin{figure*}[h!]
\centering
\includegraphics[width=1\textwidth]{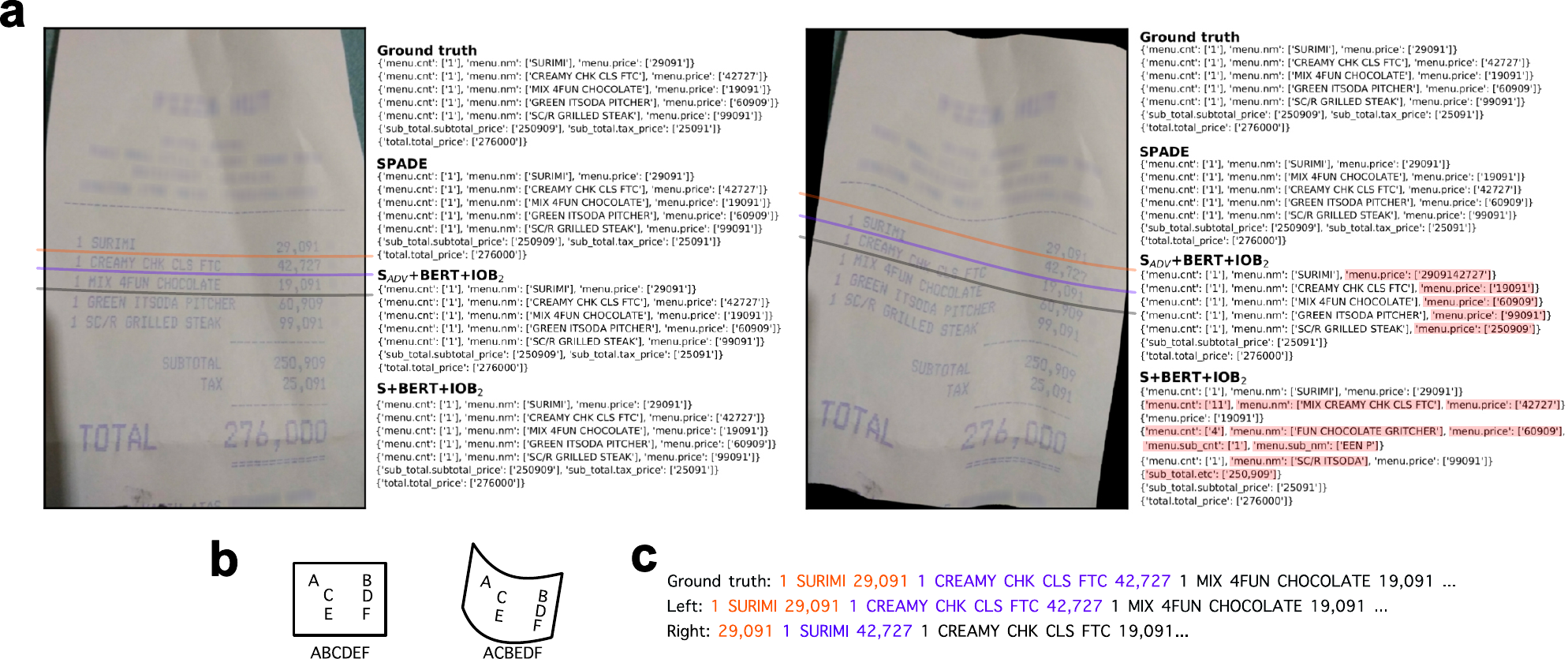}
\caption{Examples from \cord\ (left) and \cordd (right) dev sets. (a) Parses are shown in grouped key-value format with the errors in red. (b) The illustration of serialization error. (c) The input tokens serialized by $S_{\text{adv}}$.
}
\label{fig_distortion}
\vspace{-4 mm}
\end{figure*}

The main focus of \ours\ is to handle the two challenging factors of semi-structured document information extraction---complex spatial relationships and highly structured information---in a generalizable way. 
We first show that our model can handle hierarchical structure in documents by evaluating the model on two datasets \cord~\citep{park2019cord} and \receipti\, that consist of (Indonesian) receipt images.
We then show \ours\ can perform well on tasks that require modeling the complex spatial relationship in documents by reporting the performance on name card IE where the spatial layout is more complex than receipts.
Then the evaluation on the invoice dataset shows the advantage of \ours\ when both of the two challenging factors are simultaneously present.
Finally, we show that \ours\ can handle even more types of documents by evaluating the model on a form understanding dataset, FUNSD \citep{jaume2019funsd}. 
Table~\ref{tbl_results_f1} summarizes the performance of several baseline models and \ours\ in various semi-structured document information extraction tasks.

\begingroup
\setlength{\tabcolsep}{1.3pt} 
\renewcommand{\arraystretch}{1} 
\begin{table}[th]
\centering
\begin{threeparttable}
\captionsetup{font=small} 
  \caption{
  Parse prediction accuracy. 
  The datasets are referred by their abbreviations in Table. \ref{tbl_datasets}.
  $\Delta F_1$ indicates the difference between \ours\ (2nd row) and \pot\ (4th row).
  }
  \scriptsize
  \centering

  \label{tbl_results_f1}
  \centering
  \begin{tabular}{lllllcccccc}

    \toprule
    \multicolumn{1}{c}{} &                  
    \multicolumn{4}{c}{test (+oracle$^\dagger$)} &
    \multicolumn{6}{c}{test} 
    \\
    \cmidrule(r){2-5}
    \cmidrule(r){6-11}
    Model & \cordabb & \receiptiabb &\ncabb & \invoiceabb  & \cordabb & \corddlabb &\corddabb & \receiptiabb & \ncabb & \invoiceabb  \\
    \midrule
    
    \oursb  & 91.5 &92.7 & 94.0 & 87.4 & 87.4 & 86.1 &82.6 & 88.5&91.1 & 84.5  \\
    \ours  & 92.5 & 93.3 & 94.3 & 88.1 &88.2 & 87.4 & 83.1& 89.1 & 91.6 & 85.0  \\
    \potbb  & 92.4$^*$ & 93.3$^*$ & - &-& 90.1& 74.0 & 52.0 & 88.1  & - & - \\
    \pot & 92.5$^*$ & 93.4$^*$ &94.4$^*$ & 84.9$^*$ & 90.1& 85.4 & 64.8 & 89.3 &90.5 & 83.1 \\
    \midrule
    $\Delta F_1$  & 0 & -0.1 & -0.1 & +3.2 & -1.9 & +2.0 & +18.3 & -0.2 &+1.1& +1.9  \\
    \midrule
    
    \ubsiob & 58.1 &65.4 & 100 & 83.2 &- &- &- &- &- &- \\

    \bottomrule
  \end{tabular}
  \begin{tablenotes}
\scriptsize
\item[$\dagger$]{The input tokens are recognized by human annotators.}
\item[*]{The input tokens are line-grouped by human annotators.}
\end{tablenotes}
\end{threeparttable}
\vspace{-4mm}

\end{table}
\endgroup

\paragraph{Handling hierarchical structure in documents}
\cord\ consists of receipt images without creases or warping.
\ours\ initially achieves 91.5\% and 87.4\% in $F_1$ with and without the oracle (ground truth OCR results), respectively (Table \ref{tbl_results_f1}, 1st row, \cordabb). Their dependency parsing score is also shown in Table \ref{tbl_results_rel} in Appendix (1st panel, \cordabb).
To push the performance further, we notice that individual text nodes have a single incoming edge for each relation except in special documents like table (Fig. \ref{fig_paradigm}). Using this property, we integrate Tail Collision Avoidance algorithm (\caabb) that iteratively trims the tail-sharing-edges and generate new edges until the process becomes self-consistent (Section \ref{sec:graph_gen}).
$F_1$ increases by +1.0\% and +0.8\% with and without the oracle upon the integration (2nd row, \cordabb). 

\paragraph{Importance of generating hierarchical structure in receipt IE}
In receipt IE task, the inter-grouping of fields is critical due to multiple appearance of same field types such as \texttt{menu\_name} and \texttt{price} (Fig. \ref{fig_distortion}a). Without the field grouping, the maximum achievable score is 58.1 $F_1$ (Table \ref{tbl_results_f1}, 6th row, \ubsiob).
Generating hierarchical parses from the semi-structured documents is relatively new and thus the direct comparison to previous state-of-the-art methods are not feasible without considerable modification. General confidential issue related to industrial documents and multi-lingual properties of our task also hinder the comparison. In this regard, we build our own baselines consisting of the manually engineered serializer and BERT-based double IOB taggers (\potbb\footnote{\textsc{S} stands for the serializer.}). 

\paragraph{BERT-tagger}
The serializer generates pseudo-1D-text from the input tokens distributed in 2D and groups them line-by-line based on their height differences. 
BERT+\iobfg\ predicts the boundary between the fields and between the groups of the fields (see Section \ref{subsec:tag} for the detail).
In \cord, \potbb\ shows comparable performance with \ours\ with the oracle (-0.1 $F_1$) but shows +1.9 $F_1$ on the test set (2nd and 3rd rows, \cordabb).
The relatively lower score of \ours\ on the test set may originate from the small size of the training set (800, Table \ref{tbl_datasets}) as \ours\ needs to handle the text  serialization in a data-driven way. Indeed, when both models are trained using \receipti\ that consists of 9508 training examples, \ours\ outperforms by +1.0 $F_1$ on the test set (2nd and 3rd rows, \receipti). 

\paragraph{Inflexibility of tagging model in handling complex spatial relationships}
Next, we prepare \corddl\ and \cordd, which are more challenging setups where the images are warped or tilted as often seen in real-world applications (Fig. \ref{fig_distortion}).
\ours\ significantly outperforms \potbb\ (+13.4\% $F_1$ in \corddl, +31.1\% $F_1$.b in \cordd). This is due to the failure in the serialization in \potbb\ resulting in line-mixing (Fig. \ref{fig_distortion}b, c and Fig. \ref{fig_cord_examples1}, \ref{fig_cord_examples2} in Appendix). 
To understand how much improvement can be achieved through further manual engineering, we prepare \pot\ which is equipped with the advanced serializer where polynomial fitting is employed to group tokens placed on curvy line. The result shows although there is a large improvement in \corddl\ and \cordd\ task compared to \potbb, \ours\ still shows the better performance (+2.0\% in \corddl,  +18.3\% in \cordd, 1st and 4th rows). 
This shows the limitation of a serializer-based method that it cannot be easily generalized to handle document images in wild and the performance can be bottlenecked by the serialization step regardless of how advanced tagging models are.
The competent performance of \ours\ on \cordtc, a dataset generated by concatenating two receipt images from \cord\ into a single image (Fig. \ref{fig_two_column} in Appendix), further highlights the flexibility of \ours.

\paragraph{Handling documents having complex layout}
We further evaluate \ours\ on name card IE task. Unlike receipts, no inter-grouping between fields is necessary for name card IE. However, name cards often have a complex layout such as non-horizontal alignment of text or multi column even without tilting and warping (Fig. \ref{fig_paradigm}d). 
Our model achieves +1.1\% $F_1$ compared to \pot\ on the test set (Table \ref{tbl_results_f1}, \ncabb).

\paragraph{Handling documents having both hierarchical structure and complex layout}
To fully explore the capability of \ours, we further evaluate the model on invoice IE task.
Typical invoices have a hierarchical structure where some fields need to be grouped together, such as \texttt{item\_name}, \texttt{count}, and \texttt{price} that correspond to one same item.
In addition, invoices also have a relatively complex layout, having multiple tables or columns.
\ours\ achieves +1.9 $F_1$ compared to \pot\ (Table. \ref{tbl_results_f1}, \invoiceabb).

\paragraph{Handling general documents}
In order to see if \ours\ can handle more general kinds of documents, we use the FUNSD form understanding dataset \citep{jaume2019funsd} where document IE is performed under a more abstract setting by finding general key-value pairs and their inter-grouping (Section \ref{sec:dataset_detail_funsd}). 
The performance is measured on two OCR-independent subtasks \citep{jaume2019funsd}: (1) ``entity-labeling (ELB)'' which predicts the information category of the serialized words, and (2) ``entity-linking (ELK)'' which measures the score for key-value pair link prediction.
The evaluation reveals that \ours\ achieves the state of the art on ELK, outperforming the previous baseline by 37.3\% $F_1$ (Table \ref{tbl_results_funsd}, rightmost column).
In ELB, \ours\ achieves +11.5\% $F_1$ absolute improvement with respect to BERT-Base Tagger. Both models use BERT-Base as a backbone. 
Although the $F_1$ scores of LayoutLM are higher than our model, their contributions are orthogonal to ours since they focus on making a better pretrained model. Also, it cannot perform ELK. 
We emphasize that \ours\ solves the three subtasks--ELB, ELK, and word serialization--simultaneously, while other tagger models need to use the perfectly serialized input text and solve only entity labeling. 
The stable performance of \ours\ over randomly rotated documents (ELB-R) or shuffled tokens (ELB-S) supports this highlighting the merit of the serializer-free architecture. 
\begingroup
\setlength{\tabcolsep}{1pt} 
\renewcommand{\arraystretch}{0.9} 
\begin{table}[h]
\centering
\begin{threeparttable}
\captionsetup{font=small} 
  \caption{
  $F_1$ scores for two FUNSD subtasks: entity labeling (ELB, ELB-R, and ELB-S) and entity linking (ELK). ``Need \texttt{S}'' means the input tokens should be serialized. ``\# of \texttt{D}'' indicates the number of documents used for layout pretraining.
  }
  
  \scriptsize

  \label{tbl_results_funsd}
  \centering
  \begin{tabular}{l|cc|lll|c}
    \toprule
     Model &Need \texttt{S}& \# of \texttt{D} & ELB & ELB-R & ELB-S & ELK   \\ 
    \midrule
    Baseline$^a$ & $\circ$ &0& 57 & - & - & 4 \\ 
    BERT-Base Tagger$^*$ &$\circ$& 0& 60.1 & 43.9 (-16.2) & 42.5 (-17.6) & - \\ 
    BERT-Large Tagger$^*$&$\circ$& 0 & 64.6 & 47.6 (-17.0) & 42.7 (-21.9) & - \\ 
    LayoutLM-Base Tagger$^b$&$\circ$& 500K & 69.9 & -  & -  & - \\ 
    LayoutLM-Base Tagger$^*$&$\circ$& 11M & \textbf{78.9} & \textbf{72.5} (-6.4) & 70.2 (-8.7)  & - \\ 
    \ours$^\dagger$ &$\times$ & 0 &   71.6 & 70.5 (-1.1) & \textbf{72.0} (+0.4)$^\$$ & 41.3 \\
    \bottomrule
  \end{tabular}
  \begin{tablenotes}
\scriptsize
\item[]{
$^a$\citet{jaume2019funsd}.~~~ $^b$ From \citet{xu2019_layoutLM}.
\\ $^*$ The source code from  https://github.com/microsoft/unilm/tree/master/layoutlm.
\\ $^\$$ The separation of long input text ($>512$) into multiple independent inputs introduces small difference in $F_1$.
\\$\dagger$Five encoder layers are used for computational efficiency.
}
\end{tablenotes}
\vspace{-4mm}
\end{threeparttable}

\end{table}
\endgroup

\paragraph{Ablation study}
We probe the role of each component of \ours\ via ablation study (Table \ref{tbl_abl}). 
The performance drops dramatically upon the removal of the relative coordinate information of tokens in the self-attention layer, highlighting its importance in the serializer-free encoder (2nd row).
When the absolute coordinates are used in the input instead of the relative coordinates, $F_1$ drops by 6.9\% (3rd row).
Finally, 2.6\% drop in $F_1$ is observed upon the removal of the data augmentation during training (4th row).

\begingroup
\setlength{\tabcolsep}{8pt} 
\renewcommand{\arraystretch}{0.9} 
\begin{table}[h]
\captionsetup{font=small} 
  \caption{
  Ablation study on \cord\ dataset. 
  }
  \scriptsize

  \label{tbl_abl}
  \centering
  \begin{threeparttable}
\begin{tabular}{ll}
    \toprule
    Model &$F_1$ \\
    \midrule
    \ourss & 84.5 \\ 
    ~~~~ (-) relative coordinate  & 10.5 (-74.0)\\ 
    ~~~~ (-) relative coordinate (+) absolute coordinate &78.6 (-6.9) \\
    ~~~~ (-) data augmentation & 81.9 (-2.6) \\
    \bottomrule
  \end{tabular}
\begin{tablenotes}
\scriptsize
\item[$\dagger$] {Five encoder layers are used for computational efficiency.}
\end{tablenotes}
\end{threeparttable}
\end{table}
\endgroup

\section{Conclusion}
We present \ours, a spatial dependency parser that can extract highly structured information from documents that have complex layouts.
By formulating document IE as a spatial dependency graph construction problem, we provide a powerful unified framework that can extract hierarchical information without feature engineering.
We empirically demonstrate the effectiveness of our model over various real-world documents---receipts, name cards, and invoices---and in a popular form understanding task.

\section*{Acknowledgments}
We thank Geewook Kim for the critical comments on the manuscript and Teakgyu Hong and Sungrae Park for the helpful discussions on FUNSD experiments.

\newpage
\bibliographystyle{acl_natbib}
\bibliography{post_ocr_parsing2}


\newpage
\appendix
\onecolumn
\section{Appendices}
\label{sec:appendix}

\subsection{Dataset} \label{sec:dataset_detail}
\subsubsection{Dataset collection}
The internal datasets \receipti, \nc\ and \invoice\ are annotated by the crowd through an in-house web application following \citep{park2019cord,hwang2019pot}.
First, each text segment is labeled (bounding box and the characters inside) for the OCR task.
The text segments are further grouped according to their field types by the crowds. 
For \receipti\ and \invoice, additional group-ids are annotated to each field for inter-grouping of them.
The text segments placed on the same line are also annotated through row-ids.
For quality assurance, the labeled documents are cross-inspected by the crowds.

\subsubsection{\cord, \corddl, \cordd, and \cordtc\ for receipt IE}
\cord\ and their variant consist of 30 information categories such as \texttt{menu\_name}, \texttt{count}, \texttt{unit\_price}, \texttt{price}, and \texttt{total\_price} (Table \ref{tbl_datasets_full}). The fields are further grouped and forms the information layer at a higher level.

\subsubsection{\receipti\ for receipt IE}
\receipti\ is similar to \cord\ but includes more diverse information categories (50) such as \texttt{store\_name}, \texttt{store\_address}, and \texttt{payment\_time} (Table \ref{tbl_datasets_full}).

\subsubsection{\nc\ for name card IE}
\nc\ consists of 12 field types, including \texttt{name}, \texttt{company\_name}, \texttt{position}, and \texttt{address} (Table \ref{tbl_datasets_full}). The task requires grouping and ordering of tokens for each field. Although there is only a single information layer (field), the careful handling of complex spatial relations is required due to the large degree of freedom in the layout.

\subsubsection{\invoice\ for invoice IE}
\invoice\ consists of 62 information categories such as \texttt{item\_name}, \texttt{count, price\_with\_tax, item price\_without\_tax, total\_price, invoice\_number, invoice\_date, vendor\_name}, and \texttt{vendor\_address} (Table \ref{tbl_datasets_full}). 
Similar to receipts, their hierarchical information is represented via inter-field grouping.

\subsubsection{FUNSD for general form understanding} \label{sec:dataset_detail_funsd}
FUNSD form understanding task consists of two sub tasks: entity labeling (ELB) and entity linking (ELK). In ELB, tokens are classifed into one of four fields--header, question, answer, and other--while doing serialization of tokens within each field.  Both subtasks assume that the input tokens are perfectly serialized with no OCR error. To emphasize the importance of correct serialization in the real-world, we prepare two variant of ELB tasks: ELB-R and ELB-S. In ELB-R, the whole documents are randomly rotated by a degree of -20\degree--20\degree\ and the input tokens are serialized using rotated y-coordinates. In ELB-S task, the input tokens are randomly shuffled. In both tasks, the relative order of the input tokens within each field remain unchanged.
In ELK task, tokens are linked based on their key-value relations (inter-grouping between fields). For example, each ``header'' is linked to the corresponding ``question'', and ``question'' is paired with the corresponding ``answer''.

\begingroup
\setlength{\tabcolsep}{3pt} 
\renewcommand{\arraystretch}{0.50} 
\begin{table}[h]
\centering
\begin{threeparttable}

\captionsetup{font=small} 
  \caption{
  The representative fields of the datasets. 
  }
  \label{tbl_datasets_full}
  \tiny

  \centering
  \begin{tabular}{cc}


    \toprule
     Dataset & representative fields and their numbers  \\
     \midrule
     \midrule
     \cord,\corddl, \cordd,\cordtc &
     \makecell{
     \texttt{menu\_name} (2572), \texttt{count} (2357), \texttt{unit\_price} (737), 
     \texttt{price} (2559), 
     \texttt{total\_price} (974) 
     }
     \\
     \midrule
     \receipti & 
     \makecell{
     \texttt{menu\_name}  (28832), \texttt{munu\_count} (27132), \texttt{menu\_unitprice} (11530),\\ 
     \texttt{menu\_price} (28028), 
     \texttt{total\_price} (10284), 
     \texttt{store\_name} (9413), \texttt{payment\_time} (9817) 
     }
     \\
     \midrule
     \nc & 
     \makecell{
     \texttt{name} (25917), \texttt{company\_name} (24386), \texttt{position} (22848), \texttt{address} (26018) 
     }
     \\
     \midrule
     \invoice & 
     \makecell{
     \texttt{item\_name} (2761),
     \texttt{count} (1950),
     \texttt{price\_with\_tax}(781),
     \texttt{price\_without\_tax} (2230),
     \\
     \texttt{total\_price} (844),
     \texttt{invoice\_number} (803),
     \texttt{invoice\_date} (987),
     \texttt{vendor\_name} (993),
     \texttt{vendor\_address} (993),
     }
     \\
     \midrule
     FUNSD$^{a}$ & \texttt{header} (563), \texttt{question} (4343), \texttt{answer} (3623), \texttt{other} (1214)  \\
    \bottomrule
  \end{tabular}
  \begin{tablenotes}
\scriptsize
\item[a]{From \citep{jaume2019funsd}.} 
\end{tablenotes}
\end{threeparttable}
\vspace{-4mm}
\end{table}
\endgroup

\subsection{Evaluation metric} \label{sec:eval_detail}
During calculation of $F_1$ for parses, the difference between prediction and ground truth is not counted in \texttt{store\_name}, \texttt{menu\_name}, and \texttt{item\_name} fields in receipt and invoice when the edit distance (ED) is less then 2 or when the ED/gt-string-length $\leq$ 0.4. Also, in Japanese documents, white spaces are ignored.

In the FUNSD form understanding task, we measure entity labeling (ELB) and entity linking (ELK) scores following \citep{jaume2019funsd}. 
ELB measures the field classification accuracy of already ``perfectly'' serialized tokens of each field (words group), whereas ELK measures the inter-grouping accuracy between word groups.
As \ours\ does both the serialization of the fields and grouping between fields simultaneously, we do not feed the serialized tokens into \ours\ but only use the oracle information to indicate the first text node of each field from the predicted graph. These text nodes effectively represent entire fields and are used for the evaluation.

\subsection{The score for the dependency relation prediction}
\begingroup
\setlength{\tabcolsep}{3pt} 
\renewcommand{\arraystretch}{0.50} 
\begin{table}[h]
\centering
\begin{threeparttable}
\captionsetup{font=small} 
  \caption{
  The score for the dependency relation prediction. \texttt{s} and \texttt{g} stand for \texttt{rel-s} and \texttt{rel-g}.
  }
  \scriptsize
    \tiny
  \label{tbl_results_rel}
  \centering
  \begin{tabular}{c@{\extracolsep{\fill}}cccccccccccccccc}
    \toprule
    \multicolumn{2}{c}{} &                  
    \multicolumn{5}{c}{Precision} &
    \multicolumn{5}{c}{Recall} &
    \multicolumn{5}{c}{$F_1$} 
    \\
    \cmidrule(r){3-7}
    \cmidrule(r){8-12}
    \cmidrule(r){13-17}
    Model& \texttt{rel} & \cordabb & \receiptiabb& \ncabb & \invoiceabb & \funsdabb & \cordabb & \receiptiabb& \ncabb & \invoiceabb & \funsdabb & \cordabb  & \receiptiabb& \ncabb & \invoiceabb & \funsdabb\\
    \midrule
    \midrule
    \oursbabb & \texttt{s} & 96.4 & 97.7&90.7 & 97.4 & 60.6$^\dagger$ & 97.1  & 98.8&\textbf{92.0} & \textbf{98.3} &\textbf{ 63.7}$^\dagger$ & 96.8 &98.3 &91.3  & 97.8 &62.2$^\dagger$ \\
    \oursbabb & \texttt{g} & 87.8 & 91.1&- & 86.7 & 41.1$^\dagger$ &\textbf{90.1 }& \textbf{93.8}&- & \textbf{88.0} &\textbf{34.4}$^\dagger$ & 88.9 &92.4& -  & 87.3 & 37.4$^\dagger$ \\
    \midrule
    \oursabb & \texttt{s} & \textbf{96.8} & \textbf{97.8} &\textbf{91.9} & \textbf{97.6} & \textbf{70.4}$^\dagger$ & 97.1 &98.8 & 91.3 & 98.2 & 59.8$^\dagger$ & \textbf{96.9}&98.3 & \textbf{91.6}  & \textbf{97.9} & \textbf{64.6}$^\dagger$ \\
    
    \oursabb &\texttt{g} & \textbf{89.9}&\textbf{92.2} &- & \textbf{88.6} & \textbf{49.7}$^\dagger$ &89.2&93.1 & - &86.3 &30.5$^\dagger$ & \textbf{89.6}&\textbf{92.7} & -  & \textbf{87.4} & \textbf{37.8}$^\dagger$ \\
    \midrule
    \ubns & \texttt{s} & 100 &100 & 100 & 100 & - & 32.7 & 31.3 &57.7 & 18.8 &- & 49.3 &47.7 & 73.1 & 31.7 & - \\

    \ubns & \texttt{g} & 0  & 0& - & 0 & - & 0  & 0& - & 0 &- & 0 & 0 & -  & 0 & - \\
    \bottomrule
  \end{tabular}
  \begin{tablenotes}
\scriptsize
\item[]{
~~~$\dagger$Five encoder layers are used instead of twelve for computational efficiency.
}
\end{tablenotes}
\end{threeparttable}
\vspace{-4mm}
\end{table}
\endgroup

\begin{figure}[h]
\centering
\includegraphics[width=0.6\textwidth]{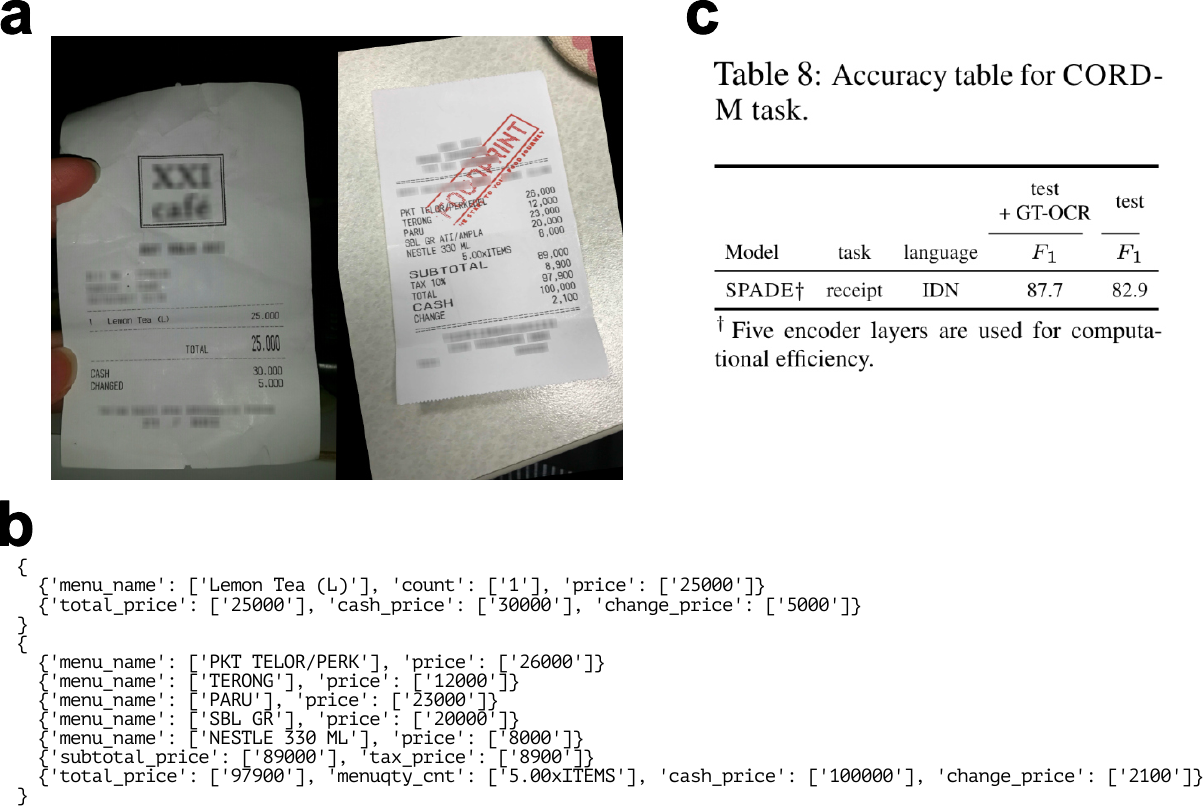}
\caption{The example of a receipt image from \cordtc\ (a), the predicted parse (b), and the accuracy table (c).
}
\label{fig_two_column}
\end{figure}

\begin{figure}[h]
\centering
\includegraphics[width=1\textwidth]{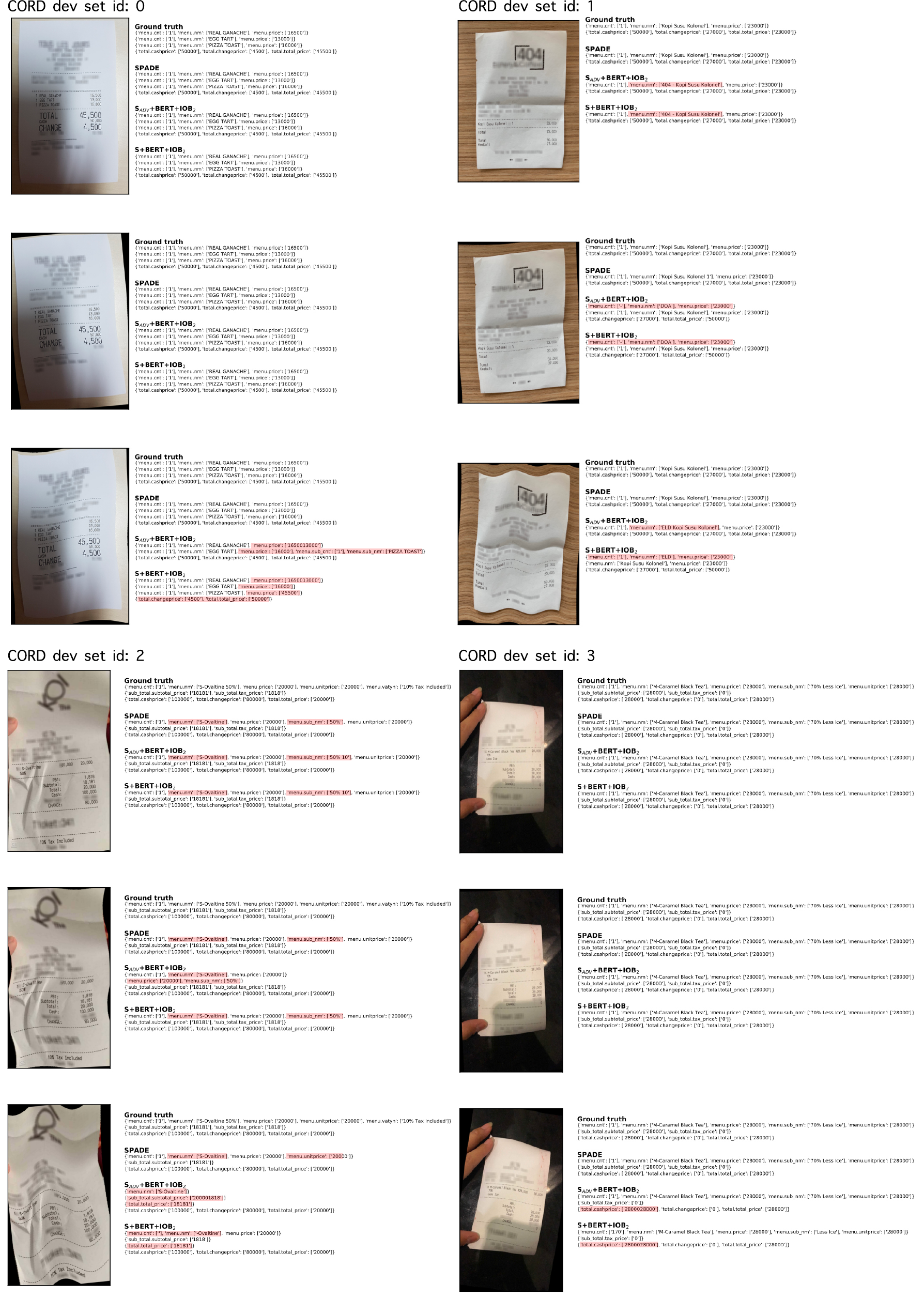}
\caption{The example from \cord, \corddl, and \cordd\ dev sets (ids 0--3).}
\label{fig_cord_examples1}
\end{figure}

\begin{figure}[h]
\centering
\includegraphics[width=1\textwidth]{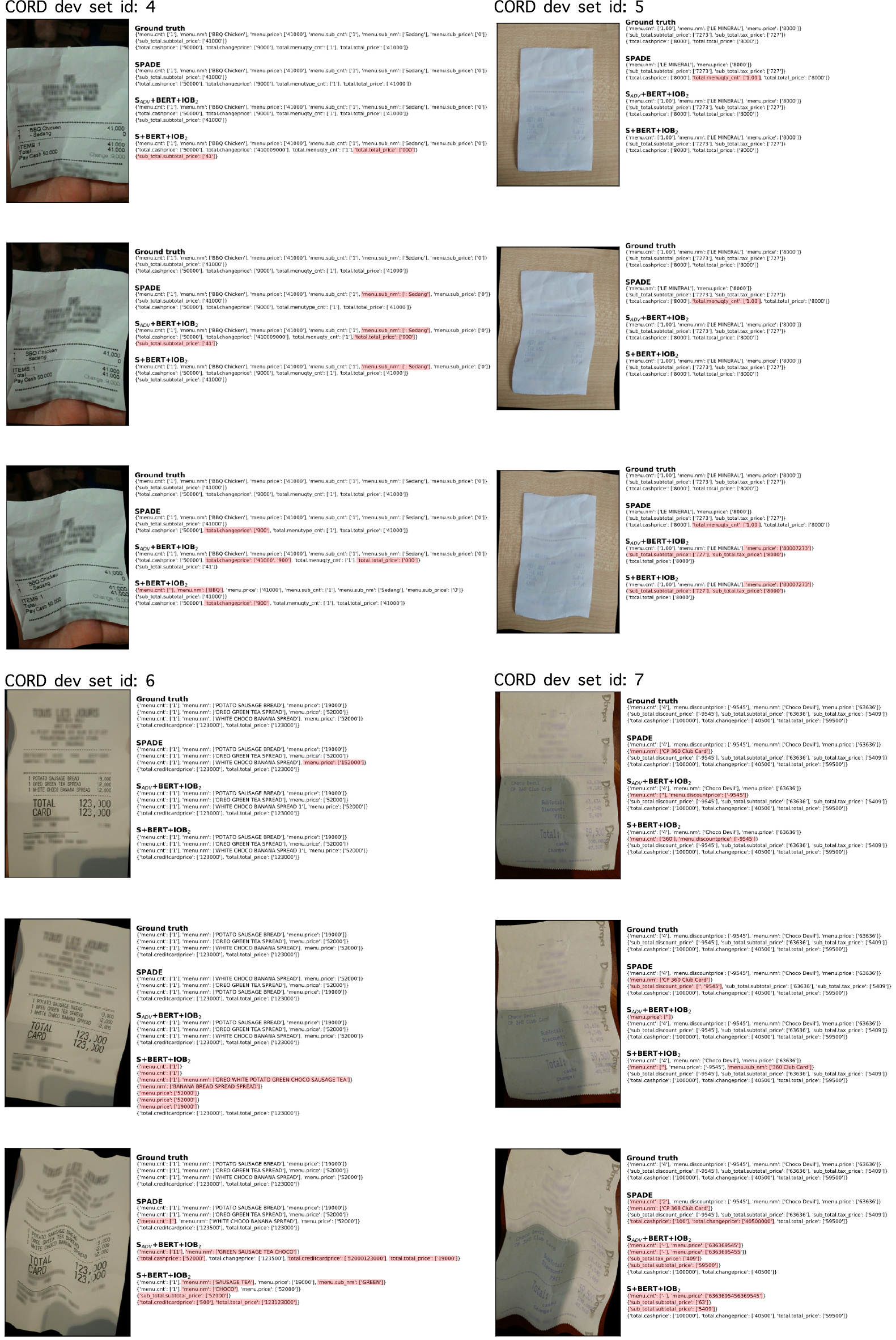}
\caption{The example from \cord, \corddl, and \cordd\ dev sets (ids 4--7).}
\label{fig_cord_examples2}
\end{figure}

\end{document}